# CHD: Consecutive Horizontal Dropout for Human Gait Feature Extraction

Chengtao Cai
College of Automation
Harbin Engineering University
Harbin, China, 150001
86045182589656
caichengtao@hrbeu.edu.cn

Yueyuan Zhou
College of Automation
Harbin Engineering University
Harbin, China, 150001
8615104555855
yueyuan_zhou@hrbeu.edu.cn

Yanming Wang
Faculty of Information Technology
Beijing University of Technology
Beijing, China, 100124
8619801355807
wangym@emails.bjut.edu.cn

## ABSTRACT

Despite gait recognition and person re-identification researches have made a lot of progress, the accuracy of identification is not high enough in some specific situations, for example, people carrying bags or changing coats. In order to alleviate above situations, we propose a simple but effective Consecutive Horizontal Dropout (CHD) method apply on human feature extraction in deep learning network to avoid overfitting. Within the CHD, we intensify the robust of deep learning network for cross-view gait recognition and person re-identification. The experiments illustrate that the rank-1 accuracy on cross-view gait recognition task has been increased about 10% from 68.0% to 78.201% and 8% from 83.545% to 91.364% in person re-identification task in wearing coat or jacket condition. In addition, 100% accuracy of NM condition was first obtained with CHD. On the benchmarks of CASIA-B, above accuracies are state-of-the-arts.

## 1. INTRODUCTION

Gait is defined as the style or manner of walking[1]. Gait Recognition identifies people according to their natural walking motion[2]. By recognize a person with a sequence is a complex mission because the gait data contains 4 dimensions with spatial and temporal information.

In order to save the ability of calculation, template-based method addressed which Generating a template image utilize a sequence of silhouette images. The most popular template is gait energy image (GEI)[3]. Template-based method reduces the occupation of computing resources, making it possible to implement re-identification based on gait information under limited hardware technologies. After template-based method, with the development of hardware technology, 3D convolutional neural network (CNN) has used for gait recognition[4, 5]. The input of 3D-CNN is 4-dimensional data of ordered gait sequence within temporal information, which retain more valuable information than template-based method and improve the accuracy of person re-identification greatly. However, 3D-CNN need much more capability of calculation and memory than the template-based method so that 3D-CNN method is high-cost both on time and economy. Based on the above two methods, the third method loses less information while avoiding 3D-CNN, which has a trend of being widely used. The successful paradigm of the third method is Gaitset[6] which achieves the highest recognition accuracy. Therefore, we study based on the Gaitset.

However, the three methods above can obtain a good accuracy on normal situations but worse accuracy on those special environments, for example people wearing coat, carrying bags, walking too fast, etc. In order to improve the robustness, many algorithms have been proposed. For example, HPM[7] and PS[8] deal with the missing body cases while Hierarchical Gaussian Descriptor[9] focus on color and textural. At the same time, we propose Horizontal Dropout (HD) committing to solving the poor robustness of gait recognition problem as well.

## 2. RELATED WORK

### 2.1 Deep Learning for Person Re-ID

The Re-ID can be considered as a matching or multi-classification mission in the computer vision[10]. The data always format as $S_s = \{S_1, S_2, ..., S_{ns}\}$, where $ns$ is the number of sequences and $s$ represent sequence number.

The s sequence concluding $nf(s)$ human images are represented as:

$$f_s^n = \{f_1^1, f_1^2, ... f_1^{nf(1)}, f_2^1, f_2^2, ..., f_2^{nf(2)}, ..., f_{ns}^{nf(s)}\}$$

With the above expression, Re-ID is divided into two types of model.

#### 2.1.1 Identification model

Identification model regards Re-ID mission as classification task. The output of model is the most possible person ID of the input data. Due to the fact that identification models make full use of the training data, more and more algorithms based on identification model achieve great accuracy, for example, Y Lin et al.[11]work, D Li et al.[12]work and Spindle[13].

#### 2.1.2 Verification model

Verification model usually need two or more input data, then compare the distance between probe and each input image to decide which whether they are same person or not. PRDC[14] is the typical verification model which purpose is to make same label distance less than disparate one by reducing the distance of same ID and aggrandizing the distance of different ID label. Inspired by PRDC, D Cheng et al.[15], A Hermans et al.[16], W Chen et al.[17]addressed improved algorithm and achieved better performance.

### 2.2 Gait Recognition

Gait recognition can base on various sensor devices such as floor sensor system[18],depth sensors[19], acceleration sensors[20-23], and even WIFI signal[24]. We only cover vision-based gait recognition which is also the most popular one. The gait recognition task always pretreats human images to silhouettes, that is different from person re-identification. Inchoate gait recognition only considered the 90° case where the person walked parallel to the camera[25, 26]. Recently, cross view gait recognition has been studied more and had some achievements [6, 27, 28].

## 2.3 Dropout

Dropout was designed by N Srivastave et al.[29] to avoid overfitting in the deep neural network training by setting neurons to zero randomly. In the next few years, many applications using dropout to improve various neural network successfully, such as N Srivastave et al.[30], GE Dahl et al.[31], Y Gai et al.[32], V Pham et al.[33] and H Wu et al.[34]. Furthermore, some people are working on improving dropout. Z Li et al.[35] proposed an efficient adaptive dropout in order to deal with the issue of evolving distribution of neurons in deep learning.

## 3. PROPOSED APPROACH

In this section we will describe the structure and implementation of our approach Consecutive Horizontal Dropout (CHD). However, at first, we designed two HD structures (see Figure 1) which are Consecutive HD (CHD) and Sporadic HD (SHD).

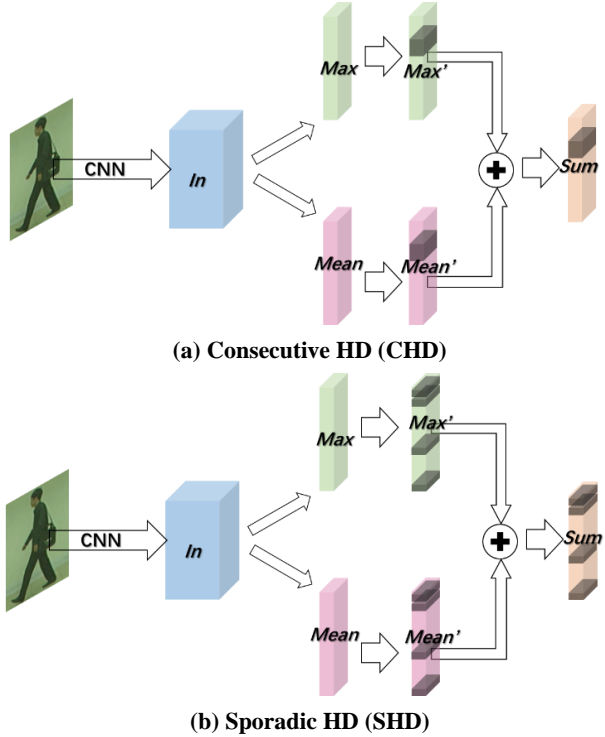

**(a) Consecutive HD (CHD)**

**(b) Sporadic HD (SHD)**

**Figure 1. The two structures of HD where CHD performs better than SHD even though SHD works as well.**

Actually, the SHD is more like traditional 3D dropout. But, the experiments in Section 4 indicate that the CHD is much better than SHD so that we recommend CHD. In theory, features of each row are not irrelevant. Before we do HD on feature, each row is a vector that represents a large view of spatial information on temporal dimension. If we split features to too many slides then the network cannot recognize the effective areas.

## 3.1 The structure of HD

The input of HD can be diverse human images such like RGB images and silhouettes. The arrow with CNN in Figure 1

represents convolutional neural network to extract features from input image as a backbone network. The features output from CNN will be used to calculate maximum and mean value on width dimension separately that reveal as mint cube and pink cube in Figure 1. After all these operations, we divide HD into two structures according to drop modality. The first is CHD as seen from (a) in Figure 1 and (b) in Figure 1 shows the second algorithm of SHD. CHD randomly zero out entire channels of a certain height while the SHD zero out entire channels on randomly rows that may not continuous. Black cube in Figure 1 express zeros. The number of drop rows is a parameter to be decided in specific application. For a clearer description, the number of drop rows will be represented as drop-number in the following sections. Finally, we add the maximum matrix and mean matrix to generate new features shown as the feldspar cube in Figure 1.

## 3.2 The Implementation of CHD

### 3.2.1 Definitions

In practical applications, multiple images are usually used as a batch for training, so that we express the input of a batch with $n$ image(s) as $Img(n)$. Analogously, the output of CNN is denoted as a matrix $In(n, c, h, w)$, where $(n, c, h, w)$ denotes the shape of matrix $In$. Then we obtain two matrices $Max(n, c, h, 1)$ and $Mean(n, c, h, 1)$ and two dropped matrices $Max'(n, c, h, 1)$ and $Mean'(n, c, h, 1)$ according to the HD structure. Finally, the sum of $Max'$ and $Mean'$ is the matrix $Sum(n, c, h, 1)$. In addition, we always regard $(n, c, h, 1)$ and $(n, c, h)$ as the same shape in this paper.

### 3.2.2 Details

To verify the effectiveness of HD, we choose GaitSet without HPM to do the experiment, however all existing CNN backbone could be used, for example, Resnet[36], VGG[37], DenseNet[38] and etc. Please note that whatever the backbone is, the output should be 4 dimensional features including the batch dimension. When we obtain the output features from CNN backbone network, we have three options to implement exactly same HD (see Figure 2). In this paper we choose option (c) in Figure 2.

First, we obtain original features $In$ from modified GaitSet shown in Figure 3 with parameters in Table 1.

**Table 1. Parameters of training.**

| | |
|---|---|
| **Learning-rate** | 0.001 |
| **Batch-size** | 11x16 |
| **GPUs** | 2 Tesla P100 |
| **margin** | 0.2 |
| **Frame-number** | 30 |
| **Date-shuffle** | False |
| **Iterations** | Depends |

Second, $Max$ and $Mean$ matrix are calculated with Formula 1 and Formula 2, where 3 denotes the third dimension count from zero.

$$Max(n, c, h) = \max(In(n, c, h, w), 3) \quad \text{Formula 1.}$$

$$Mean(n, c, h) = \text{mean}(In(n, c, h, w), 3) \quad \text{Formula 2.}$$

Third, we add $Max$ and $Mean$ matrix following the Formula 3.

$$Sum(n, c, h) = Max(n, c, h) + Mean(n, c, h)) \quad \text{Formula 3.}$$

Then we generate a number as start number randomly. Finally, we zero the entire horizontal planes from row of start number to sum of start number and drop-number.

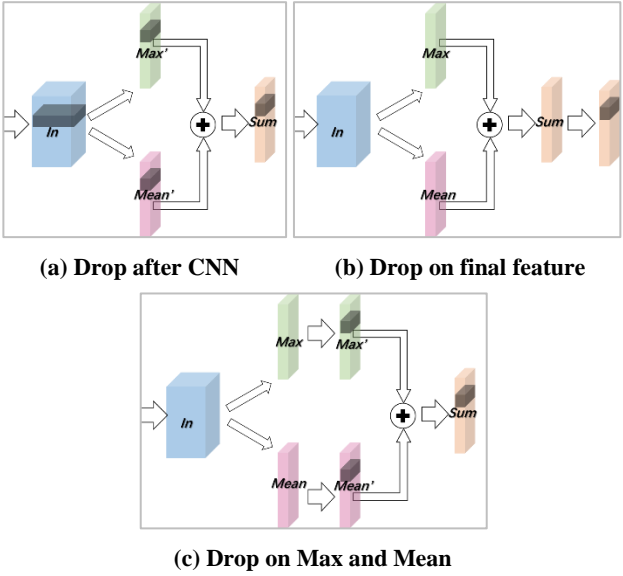

**(a) Drop after CNN**

**(b) Drop on final feature**

**(c) Drop on Max and Mean**

**Figure 2. Three options of consecutive HD implementation.**

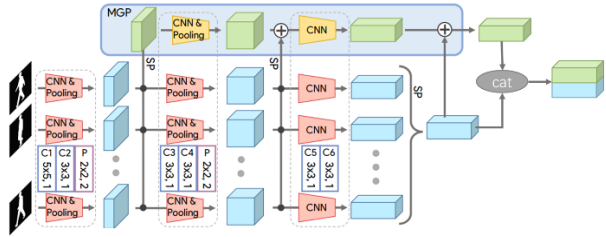

**Figure 3 The modified GaitSet structure.**

For the choice of two HD structures, CHD and SHD, we tend to be experimentally determined for specific applications. In this paper, we will display our experiment results separately on both two structures as well. The other important detail is drop-number, which has a serious impact on the results. However, we cannot give an specific number for all applications. In other words, different tasks have different best drop-number and we need to find the best value through experiments. In general, the values in area of 20 to 60 percent of height should be tested first. In this paper, we set drop-number as 2 and 16 in cross-view gait recognition task with SHD and CHD, 3 and 16 in person re-identification task with SHD and CHD。

For the case of the zero area at the bottom of the feature and the drop-number overstep the maximum value of the column in consecutive HD, the zero area will be looped to the top (see Figure 4).

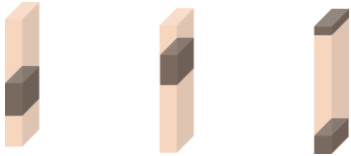

**Figure 4. The solution of zero area beyond bottom. The two left cubes represent normal random zero area in the middle of features. The right cube shows that zero area is looped to the top where the drop-number is greater than the residual rows.**

# 4. Experiments

## 4.1 Dataset

The dataset we use for training and testing our approach is CASIA-B[ 39 ] dataset which conclude 124 people over 15 thousand videos. Each person has 3 walking conditions with 10 captures which are "bg-01", "bg-02", "cl-01", "cl-02", "nm-01", "nm-02", "nm-03", "nm-04", "nm-05" and "nm-06". LT, MT and ST are 3 ways to divided training set and testing set. LT means large-sample training concluding first 74 people videos, so that the rest of 50 people's videos are used to test. Analogously, MT represent middle-sample training with 62 people and ST denotes small-sample training with 24 people in training set.

All the experiments we do using LT. For the test set, we use "nm-01", "nm-02", "nm-03", "nm-04" as gallery then employ "bg-01", "bg-02", "cl-01", "cl-02" "nm-04", "nm-05" and "nm-06" as probe.

## 4.2 Backbone

Essentially, the choice of backbone seriously affects the accuracy of experiment results. In order to prove the effect of our method, all experiment choose the newest and best network so far which is GaitSet. As a backbone network, we delete the HPP and concatenate the MGP features and main pipeline features together on the third dimension. Thus the output of the backbone is a matrix $In(n, c, h, w)$ where $h=32$ so that in our experiments drop-number is no more than 32.

**Table 2. The rank-1 identification rates [%] of cross-view gait recognition with CHD**

| Drop-number | NM | BG | CL | Average |
|---|---|---|---|---|
| 0 | 94.116 | 86.577 | 68.0 | 82.898 |
| 1 | 94.917 | 86.766 | 72.215 | 84.633 |
| 2 | **95.289** | 89.135 | 75.777 | 86.734 |
| 3 | 95.132 | 88.771 | 73.347 | 85.750 |
| 4 | 95.041 | **89.168** | 73.099 | 85.769 |
| 5 | 94.752 | 87.643 | 73.868 | 85.421 |
| 7 | 95.421 | 87.569 | 72.38 | 85.123 |
| 10 | 94.967 | 87.827 | 75.322 | 86.039 |
| 13 | 95.165 | 87.651 | 77.455 | 86.757 |
| 14 | 95.017 | 86.461 | 76.364 | 85.947 |
| 16 | 94.479 | 88.281 | **78.201** | **86.987** |
| 27 | 93.868 | 86.595 | 75.471 | 85.311 |
| 30 | 90.876 | 79.323 | 64.446 | 78.215 |
| 31 | 91.694 | 81.468 | 64.587 | 79.250 |

## 4.3 Application on Cross-View Gait Recognition

### 4.3.1 Consecutive HD

Table 2 shows the partial rank-1 accuracy results of consecutive HD applying in the gait recognition task with different drop-number. Comparing with first line without using HD, the accuracy of following lines except the last two have been improved in all three conditions (NM, BG, CL). When the drop-number is equal to 16, the accuracy in CL condition increased by over 10%. Besides, the accuracy of BG condition has been improved a lot and there is still a litter increase of NM condition as well. As can be seen from the last column, while the CL accuracy is increasing, the average accuracy is also increasing.

We aim to enhance the robustness of the CNN network to gain higher accuracy of the CL condition, so that we are inclined to choose 16. However, if improving the accuracy of NM or BG condition is the purpose, 2 or 4 drop-number should be chose according to Table 2.

### 4.3.2 Sporadic HD

Table 3 lists typical experiments results. The second to sixth lines reveal that the SHD increase the accuracy when drop-number is less than 30. Among them, when the drop-number is equal to 1, the accuracy is the best in all three conditions (NM, BG, CL).

**Table 3. The rank-1 identification rates [%] of cross-view gait recognition with SHD**

| Drop-number | NM | BG | CL | Average |
|---|---|---|---|---|
| 0 | 94.116 | 86.577 | 68.0 | 82.898 |
| 1 | **95.909** | **89.895** | **69.975** | **85.260** |
| 2 | 95.562 | 89.505 | 69.512 | 84.860 |
| 3 | 95.248 | 88.587 | 69.529 | 84.455 |
| 10 | 95.488 | 89.549 | 67.884 | 84.307 |
| 15 | 95.612 | 89.227 | 67.512 | 84.117 |
| 30 | 94.124 | 87.522 | 64.529 | 82.058 |

However, comparing with Table 2, the disadvantage of SHD is obvious. As the drop-number increases, the accuracy remains almost unchanged and even decreases slightly in the later period.

## 4.4 Application on Person Re-identification

### 4.4.1 Consecutive HD

**Table 4. The rank-1 identification rates [%] of person re-identification with CHD**

| Drop-number | NM | BG | CL | Average |
|---|---|---|---|---|
| 0 | 99.818 | 97.635 | 83.545 | 93.666 |
| 2 | **100.0** | 97.999 | 87.364 | 95.121 |
| 4 | **100.0** | **98.817** | 87.455 | 95.424 |
| 7 | 99.727 | 96.906 | 86.273 | 94.302 |
| 10 | **100.0** | 97.635 | 88.273 | 95.303 |
| 13 | **100.0** | 97.358 | 90.273 | 95.877 |
| 14 | **100.0** | 96.814 | 88.636 | 95.15 |
| 16 | **100.0** | 97.179 | **91.364** | **96.181** |
| 27 | **100.0** | 96.268 | 88.0 | 94.756 |
| 30 | 99.636 | 95.814 | 80.455 | 91.968 |

Table 4 shows the typical rank-1 accuracy results of consecutive HD applying in the person re-identification task with different drop-number. Obviously, using HD with proper drop-number increased accuracy about 8% in CL condition. In addition, the accuracies of all three conditions have been improved by consecutive HD, and it is the first time that 100% accuracy appear in the benchmark of the CASIA-B.

### 4.4.2 Sporadic HD

Table 5 demonstrate the partial rank-1 accuracy results of SHD applying in person re-identification task with different drop-number. The best result for NM condition is when drop-number is equal to 1, but BG is 2 and CL is 3. Considering that our purpose is to enhance the network recognition ability in CL condition and the average accuracy is highest when drop-number is 3, we regard 3 as the best drop-number.

**Table 5. The rank-1 identification rates [%] of person re-identification with SHD**

| Drop-number | NM | BG | CL | Average |
|---|---|---|---|---|
| 0 | 99.818 | 97.635 | 83.545 | 93.666 |
| 1 | **100.0** | 98.361 | 84.545 | 94.302 |
| 2 | 99.818 | **99.272** | 82.818 | 93.969 |
| 3 | 99.909 | 98.454 | **85.636** | **94.666** |
| 10 | 99.818 | 98.545 | 81.545 | 93.303 |
| 15 | 99.818 | 98.544 | 81.0 | 93.121 |
| 30 | 99.545 | 96.544 | 76.182 | 90.757 |

Same as cross-view gait recognition task, in person identification task, CHD performs much better than SHD.

## 5. Conclusion

The two methods we proposed, whether CHD or SHD, are both obviously effective in theory and experiments to extract human features. However, due to the fact that the rows of features are relevant, SHD will break the connection of each rows when drop-number is too large, we consider that CHD is more effective to enhance the generalization ability of deep learning network. In the future, we look forward to more applications of CHD.

## 6. ACKNOWLEDGMENTS

This work was supported by the National Natural Science Foundation of China (No. 61673129, 51674109).